# VIRTUAL MANUFACTURING
*Tools for improving Design and Production*

Philippe Dépincé, Damien Chablat (1), Peer-Oliver Woelk (2)

(1)Institut de Recherche en Communications et Cybernétique de Nantes - Unité mixte CNRS n°6597 / Ecole Centrale de Nantes, Université de Nantes, Ecole Nationale Supérieure des Techniques Industrielles et des Mines de Nantes- 1, rue de la Noë – BP 92101 – 44321 Nantes cedex 3 – France

(2)Institute of Production Engineering and Machine-Tools (IFW) – Schlosswender Strasse, 5 – D-30159 Hannover, Germany

Abstract: The research area "Virtual Manufacturing" can be defined as an integrated manufacturing environment which can enhance one or several levels of decision and control in manufacturing process. Several domains can be addressed: Product and Process Design, Process and Production Planning, Machine Tool, Robot and Manufacturing System. As automation technologies such as CAD/CAM have substantially shortened the time required to design products, Virtual Manufacturing will have a similar effect on the manufacturing phase thanks to the modelling, simulation and optimisation of the product and the processes involved in its fabrication.

## 1. INTRODUCTION

Manufacturing is an indispensable part of the economy and is the central activity that encompasses product, process, resources and plant. Nowadays products are more and more complex, processes are highly-sophisticated and use micro-technology and Mechatronic, the market demand (lot sizes) evolves rapidly so that we need a flexible and agile production. Moreover manufacturing enterprises may be widely distributed geographically and linked conceptually in terms of dependencies and material, information and knowledge flows.

In this complex and evolutive environment, industrialists must know about their processes before trying them in order to get it right the first time. To achieve this goal, the use of a virtual manufacturing environment will provide a computer-based environment to simulate individual manufacturing processes and the total manufacturing enterprise. Virtual Manufacturing systems enable early optimization of cost, quality and time drivers, achieve integrated product, process and resource design and finally achieve early consideration of producibility and affordability.

The aim of this paper is to present an updated vision of Virtual Manufacturing (VM) through different aspects. As, since 10 years, several projects and workshops have dealt with the Virtual Manufacturing thematic, we will first define the objectives and the scope of VM and the domains that are concerned. The expected technological benefits of VM will also been presented.

In a second part, we will present the socio-economic aspects of VM. This study will take into account the market penetration of several tools with respect to their maturity, the difference in term of effort and level of detail between industrial tools and academic research. Finally the expected economic benefits of VM will be presented.

The last part will describe the trends and exploitable results in machine tool industry (research and development towards the 'Virtual Machine Tool'), automotive (Digital Product Creation Process to design the product and the manufacturing process) and aerospace.

## 2.   WHAT IS VIRTUAL MANUFACTURING?

### 2.1   Virtual manufacturing definitions

The term Virtual Manufacturing is now widespread in literature but several definitions are attached to these words. First we have to define the objects that are studied. Virtual manufacturing concepts originate from machining operations and evolve in this manufacturing area. However one can now find a lot of applications in different fields such as casting, forging, sheet metalworking and robotics (mechanisms). The general idea one can find behind most definitions is that "Virtual Manufacturing is nothing but manufacturing in the computer". This short definition comprises two important notions: the process (manufacturing) and the environment (computer).

In [1, 2] VM is defined as "…manufacture of virtual products defined as an aggregation of computer-based information that … provide a representation of the properties and behaviors of an actualized product". Some researchers present VM with respect to virtual reality (VR). On one hand, in [3] VM is represented as a virtual world for manufacturing, on the other hand, one can consider virtual reality as a tool which offers visualization for VM [4].

The most comprehensive definition has been proposed by the Institute for Systems Research, University of Maryland, and discussed in [5,6]. Virtual Manufacturing is defined as "an integrated, synthetic manufacturing environment exercised to enhance all levels of decision and control" (Fig. 1).

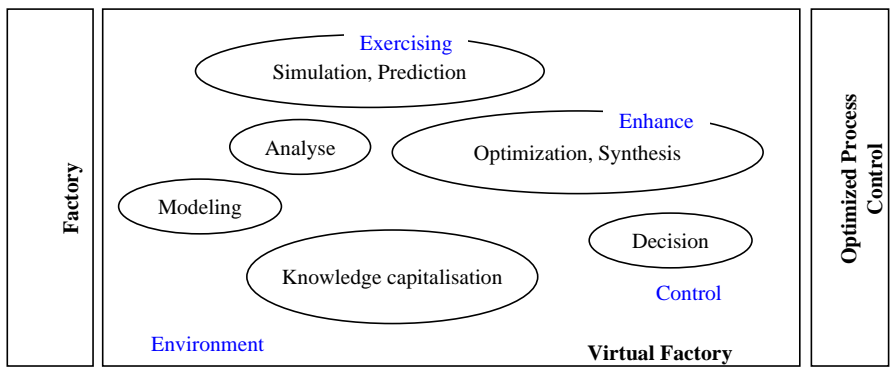

*Figure 1*: Virtual manufacturing

- **Environment:** supports the construction, provides tools, models, equipment, methodologies and organizational principles,
- **Exercising:** constructing and executing specific manufacturing simulations using the environment which can be composed of real and simulated objects, activities and processes,
- **Enhance:** increase the value, accuracy, validity,
- **Levels:** from product concept to disposal, from factory equipment to the enterprise and beyond, from material transformation to knowledge transformation,
- **Decision:** understand the impact of change (visualize, organize, identify alternatives).

A similar definition has been proposed in [7]: "Virtual Manufacturing is a system, in which the abstract prototypes of manufacturing objects, processes, activities, and principles evolve in a computer-based environment to enhance one or more attributes of the manufacturing process."

One can also define VM focusing on available methods and tools that allow a continuous, experimental depiction of production processes and equipment using digital models. Areas that are concerned are (i) product and process design, (ii) process and production planning, (iii) machine tools, robots and manufacturing system and virtual reality applications in manufacturing.

## 2.2 The scope of Virtual Manufacturing

The scope of VM can be to define the product, processes and resources within cost, weight, investment, timing and quality constraints in the context of the plant in a collaborative environment. Three paradigms are proposed in [5]:

a) Design-centered VM: provides manufacturing information to the designer during the design phase. In this case VM is the use of manufacturing-based simulations to optimize the design of product and processes for a specific manufacturing goal (DFA, quality, flexibility, …) or the use of simulations of processes to evaluate many production scenario at many levels of fidelity and scope to inform design and production decisions.

b) Production-centered VM: uses the simulation capability to modelize manufacturing processes with the purpose of allowing inexpensive, fast evaluation of many processing alternatives. From this point of view VM is the production based converse of Integrated Product Process Development (IPPD) which optimizes manufacturing processes and adds analytical production simulation to other integration and analysis technologies to allow high confidence validation of new processes and paradigms.

c) Control-centered VM: is the addition of simulations to control models and actual processes allowing for seamless simulation for optimization during the actual production cycle.

Another vision is proposed by Marinov in [7]. The activities in manufacturing include design, material selection, planning, production, quality assurance, management, marketing, …. If the scope takes into account all these activities, we can consider this system as a Virtual Production System. A VM System includes only the part of the activities which leads to a change of the product attributes (geometrical or physical characteristics, mechanical properties, …) and/or processes attributes (quality, cost, agility, …). Then, the scope is viewed in two directions: a horizontal scope along the manufacturing cycle, which involves two phases, design and production phases, and a vertical scope across the enterprise hierarchy. Within the manufacturing cycle, the design includes the part and process design and, the production phase includes part production and assembly.

We choose to define the objectives, scope and the domains concerned by the Virtual Manufacturing thanks to the 3D matrix represented in Fig. 2 which has been proposed by IWB, Munich.

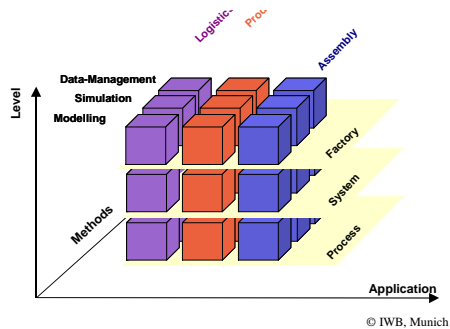

*Figure 2*: Virtual manufacturing objectives, scope and domains

The vertical plans represent the three main aspects of manufacturing today: Logistics, Productions and Assembly, which cover all aspects directly related to the manufacturing of industrial goods. The horizontal planes represent the different levels within the factory. At the lowest level (microscopic level), VM has to deal with unit operations, which include the behavior and properties of material, the models of machine tool – cutting tool – workpiece-fixture system. These models are then encapsulated to become VM cells inheriting the characteristics of the lower level plus some extra characteristics from new objects such as a virtual robot. Finally, the macroscopic level (factory level) is derived from all relevant sub-systems.

The last axis deals with the methods we can use to achieve VM systems. These methods will be discussed in the next paragraph.

## 2.3  Methods and tools used in Virtual manufacturing

Two main activities are at the core of VM. The first one is the "modeling activity" which includes determining what to model and the degree of abstraction that is needed. The second one is the ability to represent the model in a computer-based environment and to correlate to the response of the real system with a certain degree of accuracy and precision: the "simulation activity". Even if simulation tools often appears to be the core activity in VM, others research areas are relevant and necessary. One can find in [5] a classification of the technologies within the context of VM in 4 categories. A "Core" technology is a technology which is fundamental and critical to VM. The set of "Core" technologies represents what VM can do. An "Enabling" technology is necessary to build a VM system. A "Show stopper" technology is one without which a VM system cannot be built and finally a "Common" technology is one that is widely used and is important to VM.

We propose the following activities to underline methods that are necessary to achieve a VM system:

- <u>Manufacturing characterization:</u> capture, measure and analyze the variables that influence material transformation during manufacturing (representation of product/process, design by features, system behavior, …),
- <u>Modeling and representation technologies:</u> provide different kinds of models for representation, abstraction, standardization, multi-use, … All the technologies required to represent all the types of information associated with the design and fabrication of the products and the processes in such a way that the information can be shared between all software applications (Knowledge based systems, Object oriented, feature based models,…)
- <u>Visualization, environment construction technologies:</u> representation of information to the user in a way that is meaningful and easily comprehensible. It includes Virtual reality technologies, graphical user interfaces, multi context analysis and presentation,…
- <u>Verification, validation and measurement:</u> all the tools and methodologies needed to support the verification and validation of a virtual manufacturing system (metrics, decision tools,…).
- <u>Multi discipline optimization:</u> VM and simulation are usually no self-standing research disciplines, they often are used in combination with "traditional" manufacturing research.

Numerous tools are nowadays available for simulating the different levels described in Fig. 2: from the flow simulation thanks to discrete event simulation software to finite elements analysis. The results of these simulations enable companies to optimise key factors which directly affects the profitability of their manufactured products. Table 1. proposes an overview of simulation applications in manufacturing.

*Table 1.* Simulations tools

## 3. ECONOMICS AND SOCIO-ECONOMICS

| Manufacturing level | Type of simulation | Simulation targets | Level of detail |
|---|---|---|---|
| Factory / shop floor | - Flow simulation<br>- Business process simulation | - Logistic and storage<br>- Production principles<br>- Production planning and control | low |
| Manufacturing systems / manufacturing lines | - Flow simulation | - System layout<br>- Material flow<br>- Control strategies<br>- System capacity<br>- Personnel planning | Intermediate |
| Manufacturing cell / Machine tool / robot | - Flow simulation<br>- Graphical 3D kinematics simulation | - Cell layout<br>- Programming<br>- Collision test | High |
| Components | - Finite-Elements Analysis<br>- Multibody simulation<br>- Bloc simulation | - Structure (mechanical and thermal)<br>- Electronic circuits<br>- Non-linear movement dynamics | Complex |
| Manufacturing processes | - Finite-Elements analysis | - Cutting processes: surface properties, thermal effects, tool wear / life time, chip creation<br>- Metal forming processes: formfill, material flow (sheet metal), stresses, cracks | Very complex |

**FACTORS OF VIRTUAL MANUFACTURING**

### 3.1 Expected benefits

As small modifications in manufacturing can have important effects in terms of cost and quality, Virtual Manufacturing will provide manufacturers with the confidence of knowing that they can deliver quality products to market on time and within the initial budget. The expected benefits of VM are:

– from the product point of view it will reduce time-to-market, reduce the number of physical prototype models, improve quality, …: in the design phase, VM adds manufacturing information in order to allow simulation of many manufacturing alternatives: one can optimize the design of product and processes for a specific goal (assembly, lean operations, …) or evaluate many production scenarios at different levels of fidelity,

– from the production point of view it will reduce material waste, reduce cost of tooling, improve the confidence in the process, lower manufacturing cost,…: in the production phase, VM optimizes manufacturing processes including the physics level and can add analytical production simulation to other integration and analysis technologies to allow high confidence validation of new processes or paradigms. In terms of control, VM can simulate the behavior of the machine tool including the tool and part interaction (geometric and physical analysis), the NC controller (motion analysis, look-ahead)…

If we consider flow simulation, object-oriented discrete events simulations allow to efficiently model, experiment and analyze facility layout and process flow. They are an aid for the determination of optimal layout and the optimization of production lines in order to accommodate different order sizes and product mixes.

The existence of graphical-3D kinematics simulation are used for the design, evaluation and off-line programming of work-cells with the simulation of true controller of robot and allows mixed environment composed of virtual and real machines.

The finite element analysis tool is widespread and as a powerful engineering design tool it enables companies to simulate all kind of fabrication and to test them in a realistic manner. In combination with optimization tool, it can be used for decision-making. It allows reducing the number of prototypes as virtual prototype as cheaper than building physical models. It reduces the cost of tooling and improves the quality, …

VM and simulation change the procedure of product and process development. Prototyping will change to virtual prototyping so that the first real prototype will be nearly ready for production. This is intended to reduce time and cost for any industrial product. Virtual manufacturing will contribute to the following benefits [5]:

1. **Quality:** Design For Manufacturing and higher quality of the tools and work instructions available to support production;
2. **Shorter cycle time:** increase the ability to go directly into production without false starts;
3. **Producibility:** Optimise the design of the manufacturing system in coordination with the product design; first article production that is trouble-free, high quality, involves no reworks and meets requirements.
4. **Flexibility:** Execute product changeovers rapidly, mix production of different products, return to producing previously shelved products;
5. **Responsiveness:** respond to customer "what-ifs" about the impact of various funding profiles and delivery schedule with improved accuracy and timeless,
6. **Customer relations:** improved relations through the increased participation of the customer in the Integrated Product Process Development process.

## 3.2 Economic aspects

It is important to understand the difference between academic research and industrial tools in term of economic aspects. The shape of the face in the diagram presented in Fig. 3 [9], is defined by two curves:
– "effort against level of detail" where "level of detail" refers to the accuracy of the model of simulation (the number of elements in the mesh of a FEM model or the fact if only static forces are taken into account for a simulation, …
– "effort against development in time" is a type of time axis and refers to future progress and technological developments (*e.g.* more powerful computers or improved VR equipment).

Universities develop new technologies focusing on technology itself. Researchers do not care how long the simulation will need to calculate the results and they not only develop the simulation but they need to develop the tools and methods to evaluate wether the simulation is working fine and wether the results are exact. On the other hand, industrial users focus on reliability of the technology, maturity economic aspects (referring to the effort axis) and on the integration of these techniques within existing information technology systems of the companies (*e.g.* existing CAD-CAM systems, …). To our mind, Virtual Manufacturing is, for a part of its scope, still an academic topic. But in the future, it will become easier to use these technologies and it will move in the area of industrial application and then investments will pay off. For example in the automotive and aerospace companies in the late 60's, CAD was struggling for acceptance. Now 3-D geometry is the basis of the design process. It took 35 years for CAD-CAM to evolve from a novel approach used by pioneers to an established way of doing things. During this period, hardware, software, operating systems have evolved as well as education and organizations within the enterprise in order to support these new tools. Today, as shown in Fig. 4., some techniques are daily used in industry, some are mature but their uses are not widespread and some are still under development.

## 4. TRENDS AND EXPLOITABLE RESULTS

### 4.1 Machine-tool

The trend in the machine tool manufacturers sector concerning Virtual Manufacturing is research and development towards the "Virtual Machine Tool".

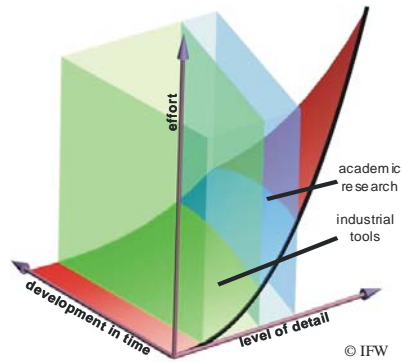

*Figure 3:* Academic research versus industrial tools

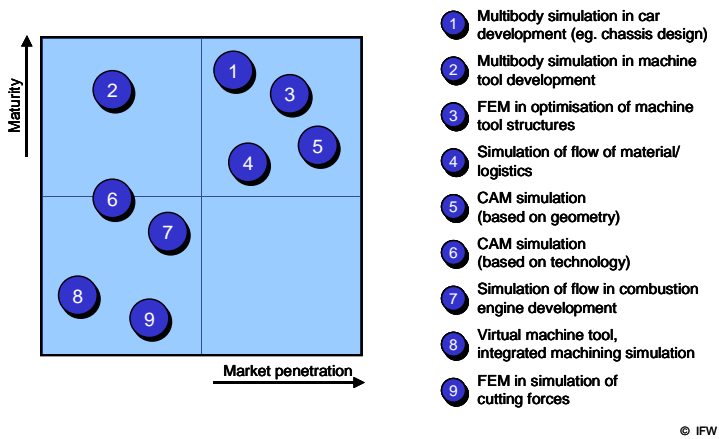

*Figure 4:* Maturity of techniques versus market penetration

The goal of the Virtual Machine Tool is to reduce time and cost for the developing of new machine tools by introducing virtual prototypes that are characterized by a comprehensive digital geometrical design, and by the simulation of (1) the stationary behavior of the machine structure, (2) the dynamic behavior of moving parts, (3) the changing of signals and state variables in electronic circuits and (4) by the simulation of the manufacturing process itself. Nowadays, the simulation activities are isolated from each other. Current research is combining different types of simulation to reflect various interdependencies, like *e.g.* elaborating the frequency response with FEA and combining this with a bloc simulation of the machine. The process simulation of forming processes is almost state-of-the-art in industry, whereas the simulation of cutting processes is an item for international research (this excludes the pure NC-program simulation, which is widely used in industry, but which does not reflect the realistic behavior of the interaction of machine tool, tool and work piece during cutting operations).

## 4.2 Automotive

In the automotive industry, the objective of the Digital Product Creation Process is to design the product and the manufacturing process digitally with full visualization and simulation for the three domains: product, process and resources. The product domain covers the design of individual part of the vehicle (including all the data throughout the product life cycle), the process domain covers the detailed planning of the manufacturing process (from the assignment of resources and optimization of workflow to process simulation). Flow simulation of factories and ware houses, 3D-kinematics simulation of manufacturing systems and robots, simulation of assembly processes with models of human operators, and FEA of parts of the automobiles are state-of-the-art.

New trends are focusing on the application of Virtual and Augmented Reality technologies. Virtual Reality technologies, like *e.g.* stereoscopic visualization via CAVE and Powerwall, are standard in product design. New developments adapt these technologies to manufacturing issues, like painting with robots. Developments in Augmented Reality focus on co-operative telework, where developers located in distributed sites manipulate a virtual work piece, which is visualized by Head Mounted Displays.

## 4.3 Aerospace

Virtual Manufacturing in aerospace industry is used in FEA to design and optimise parts, *e.g.* reduce the weight of frames by integral construction, in 3D-kinematics simulation to program automatic riveting machines and some works dealing with augmented reality to support complex assembly and service tasks (the worker sees needed information within his glasses). The simulation of human tasks with mannekins allows the definition of useful virtual environnement for assembly, maintenance and training activities [10].

## 5. CONCLUSION

As a conclusion of this paper, we can say that we have now reached a point where everyone can use VM. It appears that VM will stimulate the need to design both for manufacturability and manufacturing efficiency. Nowadays, even if there is a lot of work to do, all the pieces are in place for Virtual Manufacturing to become a standard tool for the design to manufacturing process: computer technology is widely used and accepted, the concept of virtual prototyping is widely accepted, companies need faster solutions for cost / time saving, for more accurate simulations, leading companies are already demonstrating the successful use of virtual manufacturing techniques.

Nevertheless, we have to note that there are still some drawbacks to overcome for a complete integration of VM techniques : data integrity, training, system integration. Moreover if large manufacturing enterprises have developed and applied with success VM technologies (aerospace, automotive, …, VM is a capital intensive technology and a lot of SMEs do not have the wherewithal to integrate them.

## ACKNOWLEDGEMENTS

This work has been done thanks to the EC under framework 5 "Thematic network on Manufacturing Technologies" (MANTYS), carried out under the "competitive and sustainable growth" programme (contract ref. G1RT-CT- 20001-05032) (11).